\documentclass[conference]{IEEEtran}
\IEEEoverridecommandlockouts

\def\IEEEtitletopspace{18pt}
\usepackage{cite}
\usepackage{amsmath,amssymb,amsfonts}
\usepackage{algorithmic}
\usepackage{graphicx}
\usepackage{array}
\usepackage{booktabs}
\usepackage{longtable}
\usepackage{subcaption}
\usepackage{textcomp}
\usepackage{xcolor}
\usepackage{hyperref}
\def\BibTeX{{\rm B\kern-.05em{\sc i\kern-.025em b}\kern-.08em
    T\kern-.1667em\lower.7ex\hbox{E}\kern-.125emX}}
\begin{document}

\title{Summarizing Normative Driving Behavior From Large-Scale NDS Datasets for Vehicle System Development\\
}

\author{ Gregory~Beale,
Gibran~Ali
\thanks{G. Beale and G. Ali*  are with  Virginia Tech Transportation Institute (*email: gbeale@vtti.vt.edu, gali@vtti.vt.edu)}

}

\maketitle

\begin{abstract}
This paper presents a methodology to process large-scale naturalistic driving studies (NDS) to describe the driving behavior for five vehicle metrics, including speed, speeding, lane keeping, following distance, and headway, contextualized by roadway characteristics, vehicle classes, and driver demographics. Such descriptions of normative driving behaviors can aid in the development of vehicle safety and intelligent transportation systems. The methodology is demonstrated using data from the Second Strategic Highway Research Program (SHRP 2) NDS, which includes over 34 million miles of driving across more than 3,400 drivers. Summaries of each driving metric were generated using vehicle, GPS, and forward radar data. Additionally, interactive online analytics tools were developed to visualize and compare driving behavior across groups through dynamic data selection and grouping. For example, among drivers on 65-mph roads for the SHRP 2 NDS, females aged 16–19 exceeded the speed limit by 7.5 to 15 mph slightly more often than their male counterparts, and younger drivers maintained headways under 1.5 seconds more frequently than older drivers. This work supports better vehicle systems and safer infrastructure by quantifying normative driving behaviors and offers a methodology for analyzing NDS datasets for cross group comparisons.
\end{abstract}


\section{Introduction}

Driving is inherently risky. The number of people killed in motor vehicle crashes in the United States, on average, has been increasing since 2011, and in 2023 totaled 40,901 people \cite{b1}, \cite{b2}. In 2019 the overall economic impact of the 4.5 million people injured in vehicle crashes was estimated to be \$340 billion and equated to over \$1,000 per U.S. citizen \cite{b3}.

Advanced driver assistance systems (ADAS) and automated driving systems (ADS) have shown to improve safety for many driving tasks, but most research has focused on safety-critical events \cite{b4}, \cite{b5}, \cite{b6}, \cite{b7}, \cite{b7_1}. These systems can also benefit from characterizing normal human driving behaviors, such as measures for vehicle speed, speeding, lane keeping and positioning, following distance, and headway. For ADAS and ADS to complete driving tasks comfortably and comparatively to human drivers outside of crash-imminent scenarios, the driving metrics need to be characterized and understood for normal driving conditions. These systems need to drive in reasonably predictable ways for passengers, other vehicles, and other road users to ensure occupant and roadway safety. ADAS and ADS developers and researchers would benefit from understanding normative human driving through these driving metrics, and how these typical driving behaviors change when contextualized within a range of roadway characteristics, traffic conditions, driver demographics, and vehicle classes \cite{b8}. The ability to filter or select certain normative driving behaviors based on roadway characteristics, like speed limit, or driving demographics can enable targeted group comparisons for certain driving aspects or styles and can be useful for system development.

Additionally, roadway engineers could benefit from the same comparisons for roadway design development, particularly those relevant to normative driving conditions for specific roadway conditions. Understanding current driving behavior for roadway characteristics like speed limits, road classifications, and roadway types can lead to improved and safer roadway designs in some cases \cite{b9}. Driver behavior researchers could use normative driving behaviors to establish better understanding of driving styles from the driving metrics and use those parameters to identify risky or aggressive driving from vehicle data or in other datasets or studies \cite{b10}, \cite{b11}, \cite{b12}.

Previous work has focused on targeted subsets or targeted scenarios of driving data to compare metrics such as speed, speeding, and headway values \cite{b13}, \cite{b14}, \cite{b15}, \cite{b16}, \cite{b17}, \cite{b18}, \cite{b19}. These works provide pointed insights, but do not offer complete summaries of typical driving behaviors and metrics across different roadway characteristics due to limitations in available data or scope of the research. Other works have described and cataloged lateral and longitudinal accelerations \cite{b20}, \cite{ali2023}, \cite{ali2021}, \cite{ali2024} in driving datasets, but do not address other driving metrics of interest such as speed, speeding, lane keeping, following distance, and headway or do not include large enough datasets to fully summarize normative driving behaviors. Currently, a need still exists to provide large-scale data sources of normative driving behavior contextualized by roadway and driving conditions.

To accomplish this, an underlying dataset of naturalistic driving data with a large number of drivers and high-resolution driving data is needed.  One such naturalistic driving study (NDS) is the Second Strategic Highway Research Program (SHRP 2). It includes more than 34 million miles of driving for light vehicles from around 3,500 drivers with over 5.4 million trips collected during a period of 3 years from 2010 to 2013 \cite{b21}. Participants were chosen from six locations in the U.S. and represented a sufficiently diverse group of ages, genders, races, and economic status \cite{b22}. Each trip within the SHRP 2 NDS contains high-resolution recorded data, including vehicle network variables such as vehicle speed and pedal positions; GPS location data; kinematic measurements from inertial measurement units that provide lateral and longitudinal accelerations, as well as gyroscopic information; processed radar data from a forward-mounted radar unit that provided distances and other information on other forward roadway objects; map-matched data from Here.com and Open Street Maps (OSM) \cite{b23}, \cite{b24}, \cite{b25} that provide roadway characteristic information; lane line identification data that estimates the vehicle’s lane position; and four video feeds showing the various views of the interior and exterior of the vehicle. The high-resolution vehicle data combined with the map-matching tables and large number of diverse drivers makes the SHRP 2 NDS a suitable dataset to build metrics of normative driving behavior contextualized by roadway characteristics.

This paper fulfills the need to provide measures of normative driving behavior from large-scale driving studies, contextualized by roadway characteristics, traffic conditions, driver demographics, and vehicle classes by processing the SHRP 2 NDS to create such a dataset. By doing so, this work will, one, provide a methodology for creating normative driving behaviors databases for driving metrics from NDS, and two, offer a tool for open-ended exploration of research questions for each driving metric as a \href{https://dataviz.vtti.vt.edu/SHRP2_Driving_Distributions/}{web-based interactive visualization tool}. The visualization tools created for each driving metric will allow exploration of driving behavior summary plots and statistics with dynamic data grouping, filtering, and selection across various driver demographics, vehicle classes, and roadway characteristics. Each tool is available online through a web browser for users to compare driving behaviors of different groups and investigate how these driving behaviors change as roadway characteristics or driver demographics change. This methodology can be replicated across different datasets outside of the SHRP 2 NDS, such as those that include higher percentages of ADS or other ADAS vehicle activations, to provide comparisons across various cohorts, as well.

\section{Methodology}
\subsection{Trip-level Summaries}
This work analyzed all available trips from the SHRP 2 NDS housed on the storage systems at the Virginia Tech Transportation Institute (VTTI). Each trip included vehicle data from the vehicle network or GPS, processed forward-facing radar data, associated driver demographics and vehicle class information, and map-matching data from Here and OSM containing roadway characteristics. To develop driving behavior summaries for the five driving metrics, speed, speeding, following distance, headway, and lane keeping, a large group summary dataset was created that estimated the total miles driven for each driving metric and roadway characteristic group. 

First, each individual trip was broken down into 100-millisecond time steps, and, for each timestamp, the associated vehicle and roadway variables were assigned. These included binned values for vehicle speed from the vehicle network and GPS, the speed limit and other roadway characteristics from map-matched data, and the estimated vehicle lane placement derived from lane line detection algorithms. Other vehicle parameters, such as the following distance, time-to-collision (TTC), headway, and speeding, were then calculated and binned. The headway and TTC were based on the lead vehicle that was previously identified during radar data processing. 

Bin sizes were selected to ensure adequate data resolution for each distinct driving metric while balancing the size of the data set in the subsequent grouping process. The bins needed to be large enough to aggregate similar behavior across the driving metric but small enough to capture meaningful differences in driving behavior. The vehicle speed and speeding bin width were set to 2.5 mph to capture 5- and 10-mph speed limit changes; the time headway bin width was selected to be 0.25 seconds; following distance values were binned at 10-meter intervals at ranges from 0-200 m; and vehicle lane position bin width values were set to 0.1 meters from -2.0 to 2.0 m.

Then, an estimated distance traveled was calculated based on the vehicle's current speed for the time step. After the distance metric was calculated, a group summary was performed over the entire time-labeled data to estimate the aggregate miles driven under each unique group condition for that trip. Table ~\ref{tab1} shows the composition of the data for the trip-level summaries. Note that at the end of the group summary process for each individual trip, the data no longer contained any timestamped data, just the estimated miles driven under each grouping. Each trip-level summary was appended to a common data table and the next trip was loaded to repeat the process. 

\subsection{Driving Metric Level Summaries}
\begin{figure*}[t]
    \centering
    \includegraphics[width=1.8\columnwidth,keepaspectratio]{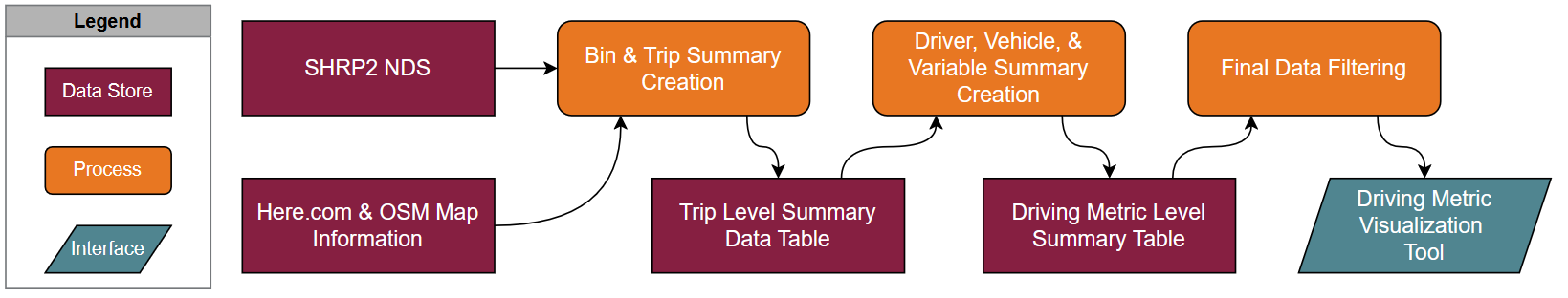}
\caption{Data flow diagram for processing the SHRP 2 NDS for the interactive visualization tool.}
\label{fig:data-flow}
\end{figure*}

Once all trip-level summaries were generated, sub-tables were created for each driving metric. This served two main functions: one, to isolate the driving metric from the larger trip-level dataset that included all the driving metric bins together; and two, reduce the size of the dataset for the visualization tools. The trip-level dataset was estimated to have a compressed size of 600-GB and include 610 million table rows after all trips were processed from the SHRP 2 NDS. Creating these driving metric sub-tables from the dataset allowed for smaller and more manageable summary tables. To accomplish this, a group summary was performed on the trip-level base table for each driving metric across the metric itself and the associated roadway characteristic values. Again, miles were aggregated for each unique grouping in the summary process to estimate the total miles driven under each group condition. Additionally, the anonymized driver demographic information (age range, gender at vehicle installation) and vehicle information (vehicle class) were joined based on the driver and vehicle information for each unique trip. Table ~\ref{tab2} shows the data table variables available for the ``speed'' group summary process. Similar tables were developed for the other selected driving metrics. After this second grouping process, the data no longer includes information from the individual trip and file IDs, but only aggregate data related to the driving metric and the driver, vehicle, and roadway characteristics. Final data filtering is applied to remove missing or extraneous data within the groupings for each driving metric. The overall data process is shown in Fig.~\ref{fig:data-flow}.
\subsection{Visualization Tool}
To investigate driving behavior trends for each driving metric, an interactive data visualization and analytics tool was developed in R using the Shiny library and deployed to the VTTI's Data Visualization web servers. A landing page was created for users to access each individual analytics tool, and is \href{https://dataviz.vtti.vt.edu/SHRP2_Driving_Distributions/}{accessible online}. The visualization tool includes two plots. The first is a step plot that shows the miles driven for each driving metric bin aggregated over all drivers. This plot can also be toggled to show the miles driven in each bin as a percentage of the total miles driven in the data set and group. The second is a box-and-whisker plot that shows the percentage of miles driven in each driving metric bin as a percentage of the total miles driven for each individual driver. This box plot gives the user a better understanding of the differences of the driving metric variables within the individual population of drivers, as opposed to the overall step plot. 

The visualization tool allows users to filter out and select specific values for certain variables in the dataset, including the speed limit, driver age range, driver gender, vehicle class, functional class, and road class characteristics. Both the step plot and box plot will dynamically update based on the user's selections, allowing a wide range of research questions to be investigated by users within each visualization tool. Any of these variables can also be selected as a faceting (i.e., separating) variable, where additional plots will be created based on the unique values within that variable. For example, the plots for a given driving metric can be generated in the visualization tool and faceted by the ``Gender'' driver demographic variable to show the differences in the driving metric for each group across the same speeding bin values. The visualization tool will plot each unique variable in the faceting group in a separate plot with a unified x-axis for direct comparisons. 

\begin{table}[htbp]
\caption{Trip-level Summary Variables for Driving Metrics}
\begin{center}
\begin{tabular}{|m{2.5cm}|m{5cm}|}
\hline
\textbf{Variable}& {\textbf{Description}} \\
\hline
File ID& Unique file ID for the trip  \\
\hline
Link ID& Here.com map-matching link ID for road segment\\
\hline
Way ID& OSM map-matching way ID for road segment\\
\hline
Functional Class& Functional class value from Here.com map-matching for the road segment\\
\hline
Speed Category& Speed category value from Here.com map-matching for the road segment\\
\hline
Road Class& Road class value from OSM map-matching for the road segment\\
\hline
Speed Limit& Speed limit (mph) from OSM and Here.com map-matching for the road segment\\
\hline
Speed& Vehicle speed bin (mph)\\
\hline
Speeding& Speeding bin (mph) relative to the speed limit\\
\hline
Distance From Center& Distance bin (m) vehicle is from center of lane\\
\hline
Headway& Time headway bin (s) between leading vehicle in same lane\\
\hline
Following Distance&  Distance bin (m) from lead vehicle in same lane\\
\hline
Miles& Aggregate miles driven for this bin combination  \\
\hline
Time& Aggregate time driven for this bin combination  \\
\hline
\end{tabular}
\label{tab1}
\end{center}
\end{table}

\begin{table}[htbp]
\caption{Driving Metric Level Summary Variables}
\begin{center}
\begin{tabular}{|m{2.5cm}|m{5cm}|}
\hline
\textbf{Variable}& {\textbf{Description}} \\
\hline
Vehicle ID& Unique vehicle ID\\
\hline
Driver ID& Unique driver ID\\
\hline
Gender& Gender of the driver at time of vehicle installation\\
\hline
Age Range& Five year age range of driver\\
\hline
Vehicle Class& Vehicle class: Car, SUV-Crossover, Truck, Minivan\\
\hline
Functional Class& Functional class value from Here.com map-matching for the road segment\\
\hline
Road Class& Road class value from OSM map-matching for the road segment\\
\hline
Speed Category& Speed category value from Here.com map-matching for the road segment\\
\hline
Speed Limit& Speed limit (mph) from OSM and Here.com map-matching for the road segment\\
\hline
Speed& Vehicle speed bin (mph)\\
\hline
Miles& Aggregate miles driven for this bin combination  \\
\hline
Time& Aggregate time driven for this bin combination  \\
\hline
\end{tabular}
\label{tab2}
\end{center}
\end{table}

The interactivity of the visualization tools is fourfold. The first is the ability to filter and facet data based on the given driver, vehicle, and roadway characteristics to make visual and statistical comparisons in the tool itself. The second is the hover information feature and dynamic plot area within the visualization web app that displays additional information on the step or box plot for each bin. This allows for user-selected data ranges for the x- and y-axis values and panning capabilities for the plot area. Thirdly, the current filtered data set and faceting variables can be downloaded from the visualization tool by the user for additional offline exploration. And fourth, the underlying driving metric level summaries for each visualization tool are \href{https://github.com/gbealeVTTI/summarizing-normative-driving-behavior-SHRP2NDS}{available online} fopr download. With this users can recreate the step and box plots generated by the visualization tool for the driving metric or inspect the underlying binned dataset. 

\section{Results}
\subsection{Large-scale NDS Data Processing}
The methodology to process 5.4 million trips in the SHRP 2 NDS produced a top-level data table with group-level summary information that included information on the selected driving metrics, roadway characteristics, unique trip IDs, and aggregate miles. Due to the number of variables in the group summary process, a large dataset with an estimated 6.1 billion rows was produced with an estimated compressed file size of 600-GB. To accommodate investigation with the interactive web-based visualization tool, the resultant driving metric level data was produced to reduce the size and introduce driver demographic and vehicle class information. Table ~\ref{tab3} shows the number of rows and estimated compressed size for the trip-level summary table and the resulting driver metric level tables. Some driving metrics require external vehicle variables to be known, which limited the total number of miles each driving metric level summary table contains. For example, the vehicle headway and following distance metrics require a lead vehicle to be present to calculate corresponding values, while the vehicle lane position metric requires the identification and processing of lane lines. If a lead vehicle is not present or lane lines cannot be detected for certain timestamps in trips, then the associated trip-level and driving metric level summary could not include aggregate miles driven data for those driving metrics. Thus, the vehicle headway, following distance, and vehicle lane position driving metric tables will have a sum of total miles driven less than the 34.4 million miles available in the SHRP 2 NDS. The speed and speeding driving metrics were not as affected by this because the vehicle speed and speed limit data were known for almost all instances of the trips.

\subsection{Interactive Visualization Tool}
The interactive visualization tools were developed and deployed to VTTI's Data Visualization web servers. Options to filter, select, and/or facet data for the roadway characteristics, driver demographics, and vehicle class are available for the step and box plots.

\begin{table}[htbp]
\caption{Data Summary Table Sizes}
\begin{center}
\begin{tabular}{|m{3cm}|m{1.8cm}|m{1.8cm}|}
\hline
\textbf{Driving Metric Table}& \textbf{Number of Rows, Millions}& \textbf{Estimated Compressed Size, GB} \\
\hline
Trip-level Summary& 6100 & 600  \\
\hline
Speed & 12.4 & 1.5  \\
\hline
Speeding & 9.5 & 1.1  \\
\hline
Headway & 15.3& 1.8  \\
\hline
Following Distance & 12.4 & 1.5  \\
\hline
Vehicle Lane Position & 3.4 & 0.4  \\
\hline
\end{tabular}
\label{tab3}
\end{center}
\end{table}

\begin{figure*}[t]
    \centering
    \subfloat[Step plot\label{fig:step-subplot}]{\includegraphics[height=5.5cm,width=0.95\columnwidth]{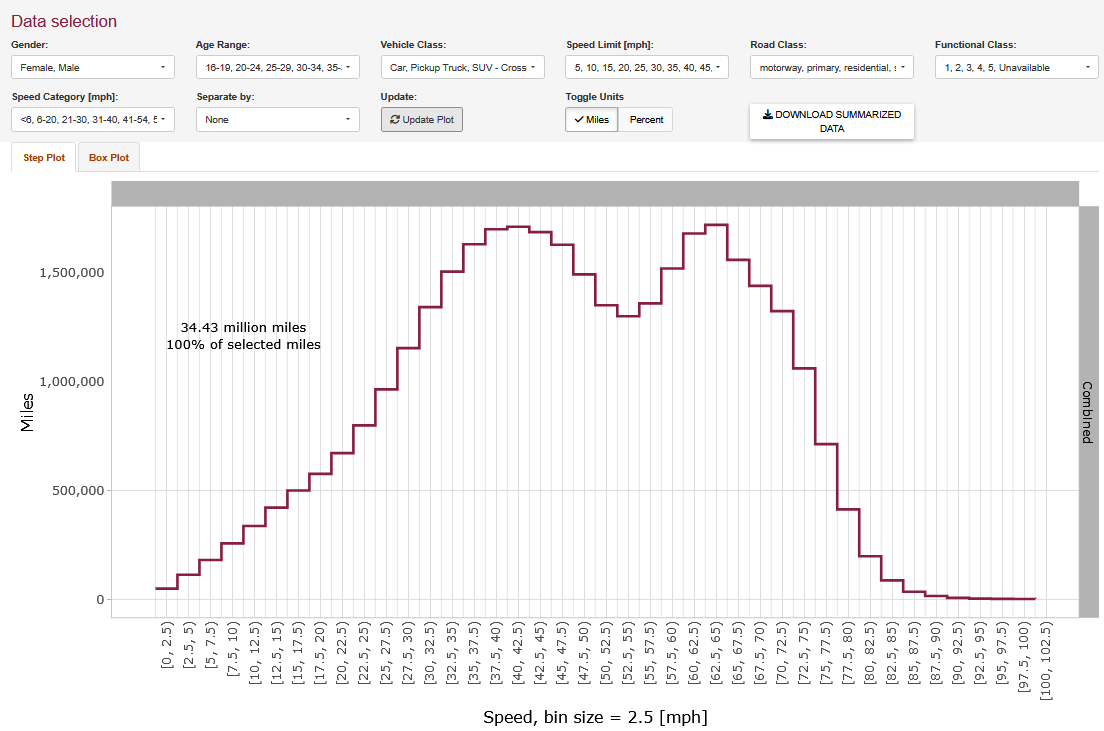}}
    \hfil
    \subfloat[Box-and-whisker plot\label{fig:box-subplot}]{\includegraphics[height=5.5cm, width=0.95\columnwidth]{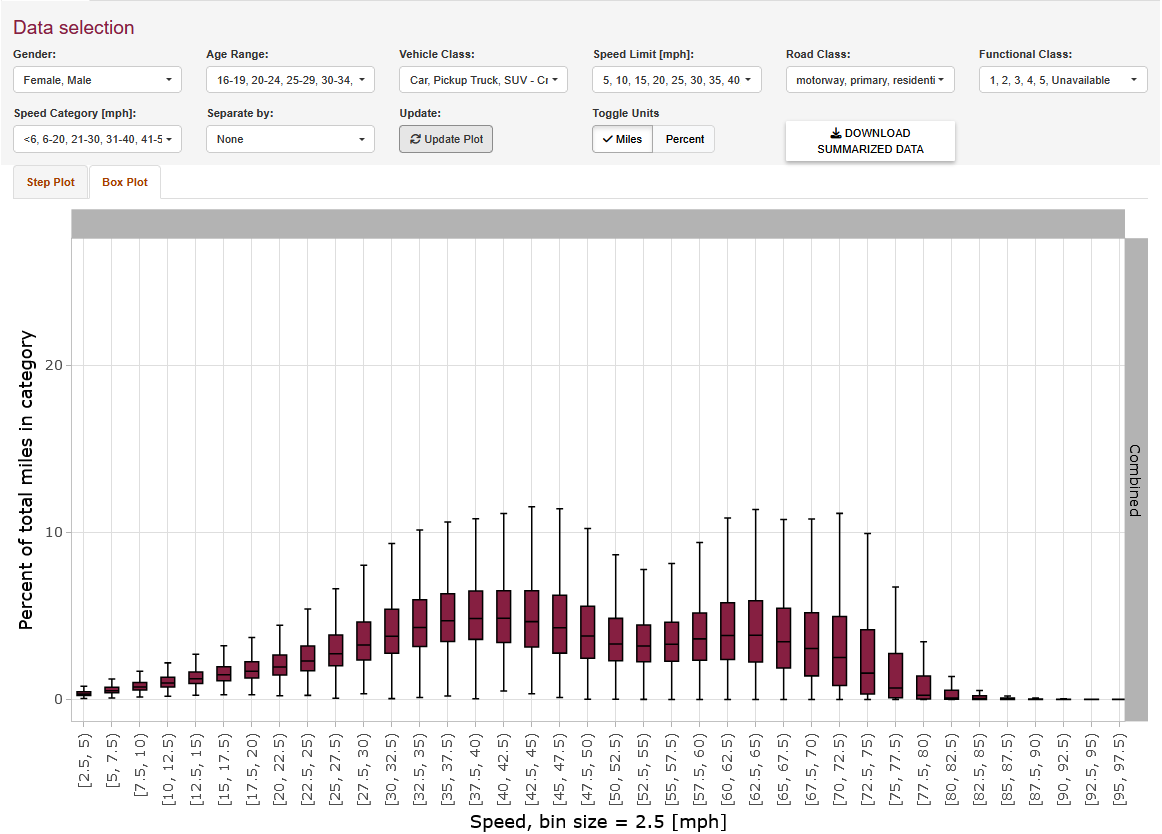}}
\caption{Interactive analytics tool step plot and box-and-whisker plot examples for the mileage driven for the given speed bins in the SHRP 2 NDS.}
\label{fig:step-plot}
\end{figure*}

% \begin{figure*}[t]
%     \centering
%     \includegraphics[width=1.2\columnwidth,keepaspectratio]{plots/BoxPlotSpeed.png}
% \caption{Interactive analytics tool box plot example for mileage driven for the given speed bins in the SHRP 2 NDS.}
% \label{fig:box-plot}
% \end{figure*}

\begin{figure*}[t]
    \centering
    \subfloat[Percentage of mileage driven for the given speeding bins for drivers aged 16-19 grouped by gender.\label{fig:speedingFacet}]{\includegraphics[height=6cm, width=0.95\columnwidth]{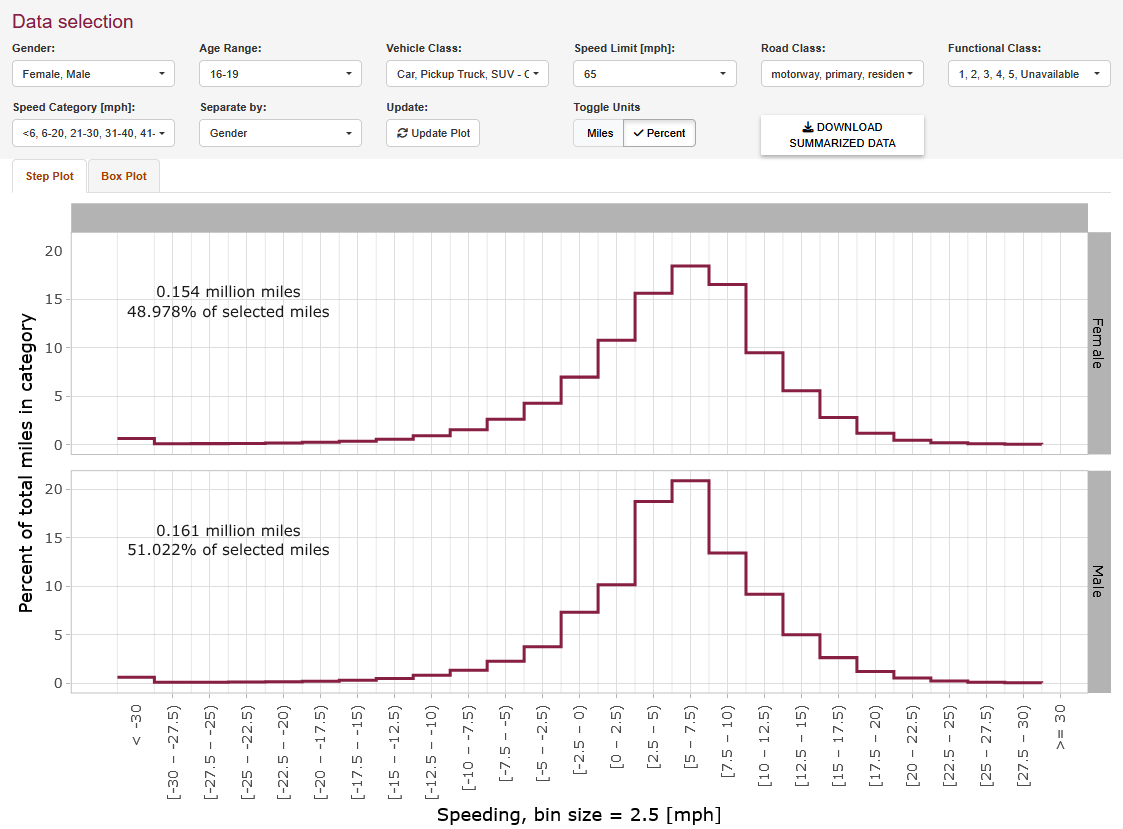}}
    \hfil
    \subfloat[Percentage of mileage driven for the given vehicle headway bins grouped by drivers aged 16-19 and 65-69.\label{fig:Headway}]{\includegraphics[height=6cm, width=0.95\columnwidth]{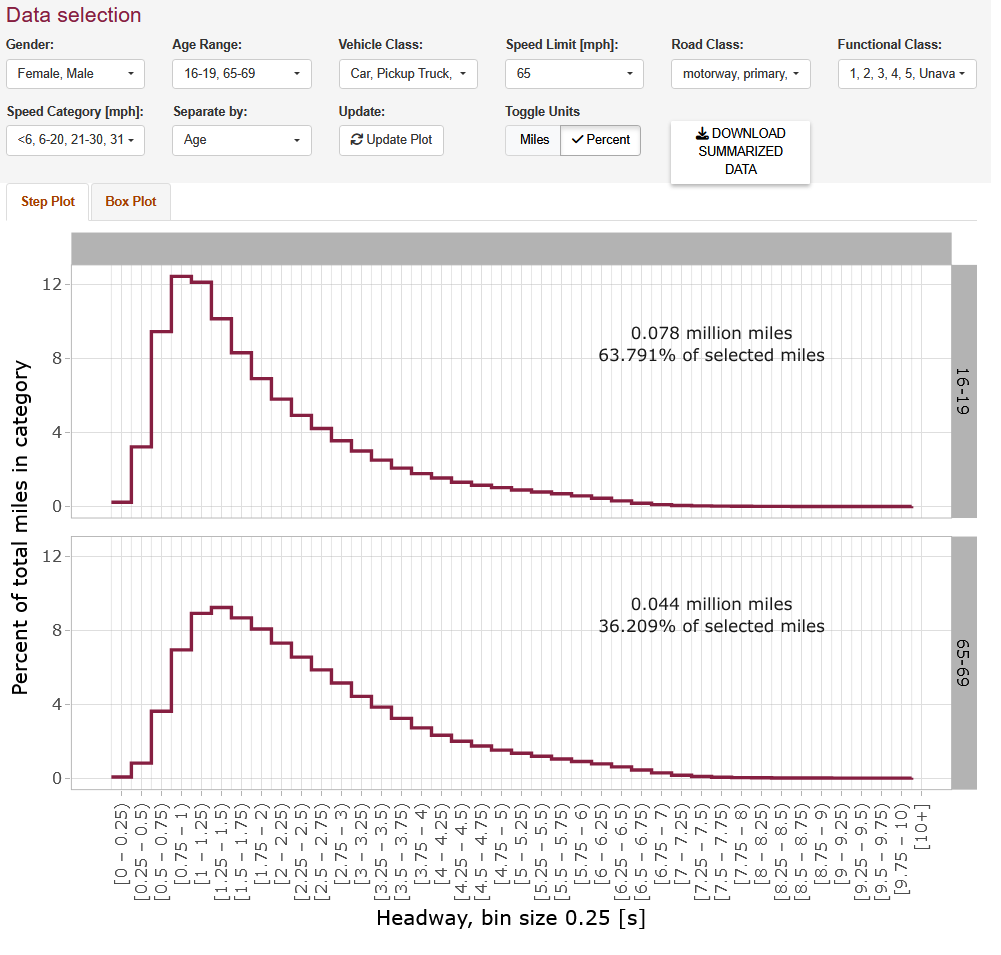}}
\caption{Interactive analytics tool step plot examples for faceting variables for the percentage of mileage driven on roads with 65 mph for the SHRP 2 NDS.}
\label{fig:speedingHeadway}
\end{figure*}

% \begin{figure*}[t]
%     \centering
%     \includegraphics[width=1.2\columnwidth,keepaspectratio]{plots/Headway65mphAgeFacet.png}
% \caption{Interactive analytics tool step plot example for the percentage of mileage driven for the given vehicle headway bins on roads with 65 mph grouped by drivers aged 16-19 and 65-69 for the SHRP 2 NDS.}
% \label{fig:Headway}
% \end{figure*}

Fig.~\ref{fig:step-subplot} shows the step plot of the aggregate miles driven for each speed bin. No data have been excluded, so all 34.4 million miles in the SHRP 2 NDS are available for exploration using the speed driving metric in this example. Fig.~\ref{fig:box-subplot} shows the box plot version of the same speed driving metric data. Similar plots are available for the remaining driving metrics. 

The data analysis used to create each plot for the driving metrics can reveal trends in driving behavior contextualized by roadway characteristics, particularly when comparing across groups. For example, a step plot of the aggregate miles driven by the speeding metric bins can be created with only the 16-19 age range and 65-mph speed limit data selected. This yields a speeding binned dataset with 315,000 total miles where the driver is between age 16-19 and is driving on roads with a 65-mph speed limit. If the gender variable is used to facet the data (see Fig.~\ref{fig:speedingFacet}), direct comparisons of the percentage of miles driven for each speeding bin between the male and female drivers of the study can be made. For the males, representing 161,000 miles of the filtered dataset, the percentage of miles driven while speeding at [0-2.5), [2.5-5.0), [5.0-7.5), [7.5-10.0), [10.0-12.5), and [12.5-15.0) mph above the 65-mph speed limit was equal to 10.17\%, 18.75\%, 20.89\%, 13.45\%, 9.19\%, and 5.01\%, respectively. For the females in the study, representing 154,000 miles of the filtered data set, the percentage of miles driven while speeding for the same speeding bins was equal to 10.79\%, 15.63\%, 18.44\%, 16.52\%, 9.49\%, and 5.57\%, respectively. 

Another example shows the comparisons between younger and older drivers using the vehicle headway driving metric. Similar step plots can be generated with data filtering performed on the speed limit variable, selecting the 65-mph speed limit, and the age range variable, selecting two age ranges: 16-19 and 65-69. If the data is faceted by the age range (see Fig.~\ref{fig:Headway}) comparisons in the vehicle headway distributions can be made across the two age groups. For the younger age group, representing 78,000 miles driven, the percentage of miles driven for the vehicle headway values of [0-0.25), [0.25-0.5), [0.5-0.75), [0.75-1.0), [1.0-1.25), [1.25-1.5), [1.5-1.75), and [1.75-2.0) seconds were 0.23\%, 3.23\%, 9.46\%, 12.45\%, 12.13\%, 10.16\%, 8.32\%, and 6.92\%, respectively. For the older grouped drivers, representing 44,000 miles driven, the percentage of miles driven for the same headway values were 0.07\%, 0.81\%, 3.62\%, 6.94\%, 8.91\%, 9.24\%, 8.67\%, and 8.07\%, respectively.

\section{Discussion}

This tool successfully catalogs the routine driving metrics for vehicle speed, speeding, following distance, headway, and lane position for the SHRP 2 NDS contextually with driver demographic information, roadway characteristics, and vehicle classification information. The web-based visualization tool allows for open-ended exploration of research questions related to the driving metrics and contextual data in NDS datasets. Dynamically filtering and visualizing data based on the large number of filtering and selection criteria helps users investigate and visually identify changes in driving distributions between groups in the driving metrics. The step plot aggregates miles driven and percentage of miles driven across groups, while the box plot lets users explore differences in the distributions of a driving metric across the population within the group (i.e., individual drivers). 

Examples of research questions that can be answered in context to the SHRP 2 NDS are differences between the speeding habits of male and female drivers of certain age groups on certain roadway conditions, such as speed limit; or the difference between the observed following distance trends of younger and older drivers for certain roadway characteristics. Direct results from the step plots developed for these two cases show that the percentage of miles for females aged 16-19 who speed [7.5-15.0] mph over the speed limit on 65-mph speed limit roads is slightly higher than the percentage of miles for males, and drivers aged 16-19 are more likely to drive with vehicle headway values less than 1.5 seconds on roads with 65-mph speed limits. The visualization tool for these driving metrics can easily be extended to answer related research questions with larger roadway criteria. Instead of selecting a single speed limit value, such as 65 mph, a range of speed limits corresponding to highway or interstate driving, such as 55, 60, 65, and 70 mph, could be selected; or a wider range of driver age groups representing senior and younger drivers could be included to compare across groups to identify trends. The large number of grouping and selection criteria can help OEMs, researchers, and vehicle developers quickly identify relevant trends in driving metric distributions from the SHRP 2 NDS.

Additionally, this work developed a template methodology to process large-scale NDS datasets into smaller, more manageable, and queryable datasets for analysis and visualization. The successful processing of the 5.4 million trips in the SHRP 2 NDS into, first, trip-level summary data, and then, second, into driving metric level summary datasets, demonstrates the data processing template. NDS datasets need to be processed first into trip-level summary with all driving metrics and roadway characteristics, and then aggregated based on driving metric and associated filtering criteria for driver, vehicle, and roadway information. The process of creating driving metric level summary data is scalable to datasets with any number of included trips. Trip-level summary data will increase size based on the number of trips within the NDS, but the driving metric level summary size only depends on the number of unique drivers and bin values for the single driving metric and the roadway characteristics. With a sufficient number of trips, all bin combinations may be populated with additional trips, only updating the aggregate miles driven value for those bin combinations. In this manner, trip-level summary data that is large, such as the 600-GB table created for the processing of the SHRP 2 NDS, can be reduced to sizes that are orders of magnitude smaller and deployable on the web for visualization tools and easier knowledge sharing. 

The methodology presented can also apply to other NDS with similar data. Creating driving behavior summaries like those presented in this work, can allow other researchers or agencies the opportunity to compare normative driving behaviors across various NDS, including those that have more ADAS or ADS instances. The template is also scalable such that the driving metric level summaries could contain multiple driving metrics in one summary table. This could allow for exploration of multivariate driving metrics together. For example, including the vehicle headway and the speeding driving metrics together could allow dynamic exploration of when drivers or groups are speeding relative to the vehicle headway across the driver demographic information, the vehicle information, and the roadway characteristics. Care should be taken to ensure driving-level metric tables stay within manageable size limits when adding multiple grouping variables for the driving metrics.

\section{Conclusions and Future Work}
This work fills the need to provide normative driving behaviors in light vehicles for vehicle speed, speeding, lane keeping, following distance, and headway for a large-scale NDS source, the SHRP 2 NDS. The methodology presented outlines a framework for processing the 5.4 million trips from over 3,400 drivers in the SHRP 2 NDS into trip-level and driving metric level tables contextualized by roadway characteristics, driver demographic information, and vehicle class information. The framework is configurable to any selection of variables of interest contained in the NDS dataset. In addition, driving metric level tables are shown to be scalable with any size of the NDS dataset based on the group summary process from the binned values. The driving metrics selected for this work correspond to those that are integral to the driving task, are typically used to categorize driving behavior, and have not been previously cataloged from the SHRP 2 NDS relative to roadway, driver, and vehicle characteristics. The interactive web-based visualization and analytics tools developed for each driving metric allow for adaptable dataset viewing, with the ability to target a large set of open-ended questions regarding the driving data. Users can quickly select, filter, or group the dataset based on roadway characteristics, such as speed limit; driver demographics, such as age range or gender; or vehicle class to visualize trends in the driving metric across groups to target research or vehicle design questions. 

Examples showed the ability to differentiate the behavior in speeding and following distance for age-separated and gender-separated groups. Results from the step plots developed for these two cases show that the percentage of miles driven for females aged 16-19 at speeds [7.5-15.0] mph over the speed limit on 65-mph speed limit roads is slightly higher than the percentage of miles driven for males, and drivers aged 16-19 are more likely to drive with vehicle headway values less than 1.5 seconds on roads with 65-mph speed limits than drivers aged 65-69. Future work considerations could include combinations of routine driving metrics into the driving metric level tables to add additional filtering and selection options for the visualization tool; or adding more grouping variable combinations to the visualization tool logic to help users compare driving behaviors of smaller groups or subgroups against larger populations within the NDS. This could enable users to compare groups such as female drivers aged 16-19 to all other drivers in gender and age groups.

It is important for safety researchers, roadway engineers, and vehicle designers to understand normative driving behaviors in order to identify risky or dangerous driving, design safer roadways and safer vehicles based on current driving behaviors, and develop more intuitive and predictable ADAS and ADS features for vehicles. This work contributes to that understanding for normative driving behavior contextualized by roadway characteristics, driver demographics, and vehicle classes derived from the large-scale SHRP 2 NDS.

\section*{Acknowledgment}
This work was funded by the \href{https://www.vtti.vt.edu/national/nstsce/index.html}{National Surface Transportation Safety Center for Excellence (NSTSCE)}.
\section*{Supplementary Data}
If viewing offline, please note the following URL to access the landing page of the interactive analytics tools, \url{https://dataviz.vtti.vt.edu/SHRP2_Driving_Distributions/} and the github page containing the underlying driving metric level tables, \url{https://github.com/gbealeVTTI/summarizing-normative-driving-behavior-SHRP2NDS}.

\vspace{12pt}

\end{document}